\documentclass{article}

\usepackage{arxiv}

\usepackage[utf8]{inputenc} 
\usepackage[T1]{fontenc}    
\usepackage[hidelinks]{hyperref}       
\usepackage{url}            
\usepackage{booktabs}       
\usepackage{amsmath}
\usepackage{float}
\usepackage{amsfonts}       
\usepackage{nicefrac}       
\usepackage{microtype}      
\usepackage{lipsum}		
\usepackage{graphicx}
\usepackage{natbib}
\usepackage{doi}
\usepackage{xurl}

\title{More is Less? A Simulation-Based Approach to Dynamic Interactions between Biases in Multimodal Models}

\date{December 23, 2024} 

\author{ 
    \href{https://orcid.org/0000-0001-6084-0317}{\includegraphics[scale=0.06]{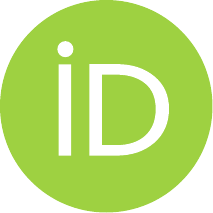}\hspace{1mm}Mounia~Drissi} \\
    Mohammed Bin Rashid School of Government\\
    Dubai, United Arab Emirates \\
    \href{mailto:mounia.drissi@mbrsg.ac.ae}{mounia.drissi@mbrsg.ac.ae}
}

\renewcommand{\shorttitle}{Dynamic Interactions between Biases in Multimodal Models}

\hypersetup{
    pdftitle={More is Less? A Simulation-Based Approach to Dynamic Interactions between Biases in Multimodal Models},
    pdfsubject={Machine Learning, Bias Analysis, Multimodal Models},
    pdfauthor={Mounia Drissi},
    pdfkeywords={bias mitigation, multimodal models, machine learning},
}

\begin{document}
\maketitle

\bigskip

\begin{abstract}
Multimodal machine learning models, such as those that combine text and image modalities, are increasingly used in critical domains including public safety, security, and healthcare. However, these systems inherit biases from their single modalities. This study proposes a systemic framework for analyzing dynamic multimodal bias interactions. Using the MMBias dataset, which encompasses categories prone to bias such as religion, nationality, and sexual orientation, this study adopts a simulation-based heuristic approach to compute bias scores for text-only, image-only, and multimodal embeddings. A framework is developed to classify bias interactions as amplification (multimodal bias exceeds both unimodal biases), mitigation (multimodal bias is lower than both), and neutrality (multimodal bias lies between unimodal biases), with proportional analyzes conducted to identify the dominant mode and dynamics in these interactions. The findings highlight that amplification (22\%) occurs when text and image biases are comparable, while mitigation (11\%) arises under the dominance of text bias, highlighting the stabilizing role of image bias. Neutral interactions (67\%) are related to a higher text bias without divergence. Conditional probabilities highlight the text's dominance in mitigation and mixed contributions in neutral and amplification cases, underscoring complex modality interplay. In doing so, the study encourages the use of this heuristic, systemic, and interpretable framework to analyze multimodal bias interactions, providing insight into how intermodal biases dynamically interact, with practical applications for multimodal modeling and  transferability to context-based datasets, all essential for developing fair and equitable AI models.
\end{abstract}

\smallskip

\twocolumn

\section{Introduction}

Multimodal models that integrate diverse data types, such as text with images, are increasingly used across critical sectors \citep{1,2,3,4,5,6,7,8,9,10,11,12,13}. For instance, to detect inappropriate conduct, content moderation, surveillance, and defense target identification rely on analyzing text, images, and videos together to provide contextually accurate decisions \citep{14,15,16}. Multimodality is also becoming relevant in analyzing public opinion \citep{17}, with effective models predicting the potential spread of false information \citep{18}. 

The biases inherent to each modality are well known \citep{19,20,21,22,23,24,25}. For example, language models are often found to be capable of reinforcing stereotypes \citep{26,27,28}, and image classifiers have the potential to enhance misrepresentation \citep{29}. In principle, multimodal models should reduce these biases through triangulation and the integration of complementary inputs, leveraging this potential to construct reliable and valid information. However, emerging research reveals that they often fail in this task; instead, they may introduce new errors with significant implications. To illustrate, in content moderation, the combination of text and image was found to amplify their biases, resulting in the disproportionate banning of targeted content \citep{15}. Multimodal fusion in military target detection, such as combining electro-optical (EO) imagery with infrared (IR) or synthetic aperture radar (SAR) data, was found to misidentify neutral targets as threats \citep{30}, potentially leading to wrongful military engagements and unjustified life losses.

Emerging research has provided key insights into bias detection and mitigation in multimodal models \citep{31,32,33,34,35,36,37,38,39,40}. Despite these efforts, there remains a significant gap in the understanding of the interplay of these biases across modalities.  For instance, in text-to-image generators like DALL-E v2, simple prompts like "a photo of a CEO" often generate non-diverse outputs, even without explicit bias in the input, suggesting bias transfer from one modality to another \citep{41}. In addition, gender bias in the visual data was found to seep into the generated text, with visually biased images producing correspondingly biased captions \citep{42,43}. These cross-modal bias examples suggest a deeper systemic issue: biases in multimodal models may not just accumulate but interact in ways that are poorly understood. Despite this complexity, most researchers -likely pushed by the rush to mitigate AI inaccuracies- remain fixated on bias mitigation strategies, with little attention paid to clarifying these underlying mechanisms. 

To address this, the study fills a critical gap in understanding the dynamic interplay between biases in multimodal models by asking whether dynamic interactions can be systemically identified and characterized. As a foundational step, this study proposes a simulation-based heuristic approach that integrates bias quantification for text and image biases with their combined effect in multimodal settings. Bias scores were simulated probabilistically using controlled random sampling to represent the varying degrees of bias for each modality. Multimodal bias is computed as a weighted combination of text and image biases, with additional noise to mimic real-world variability. The interaction effects—amplification, mitigation, or neutrality—are determined using a simple rule-based classification system by comparing multimodal bias scores to individual modality scores. This approach provides a foundational and interpretable framework for future studies to simulate and analyze dynamic multimodal bias interactions.  Such an understanding is essential for developing fair, equitable, and accurate artificial intelligence (AI) systems that minimize the multimodal risks of perpetuating harm, especially in critical applications. 

\section{Related work}

While multimodal models promise groundbreaking advancements, they also risk amplifying existing biases. However, the complex interplay between biases remains largely underexplored.

\begin{figure*}[htbp]
	\centering
	\fbox{\includegraphics[width=0.98\textwidth]{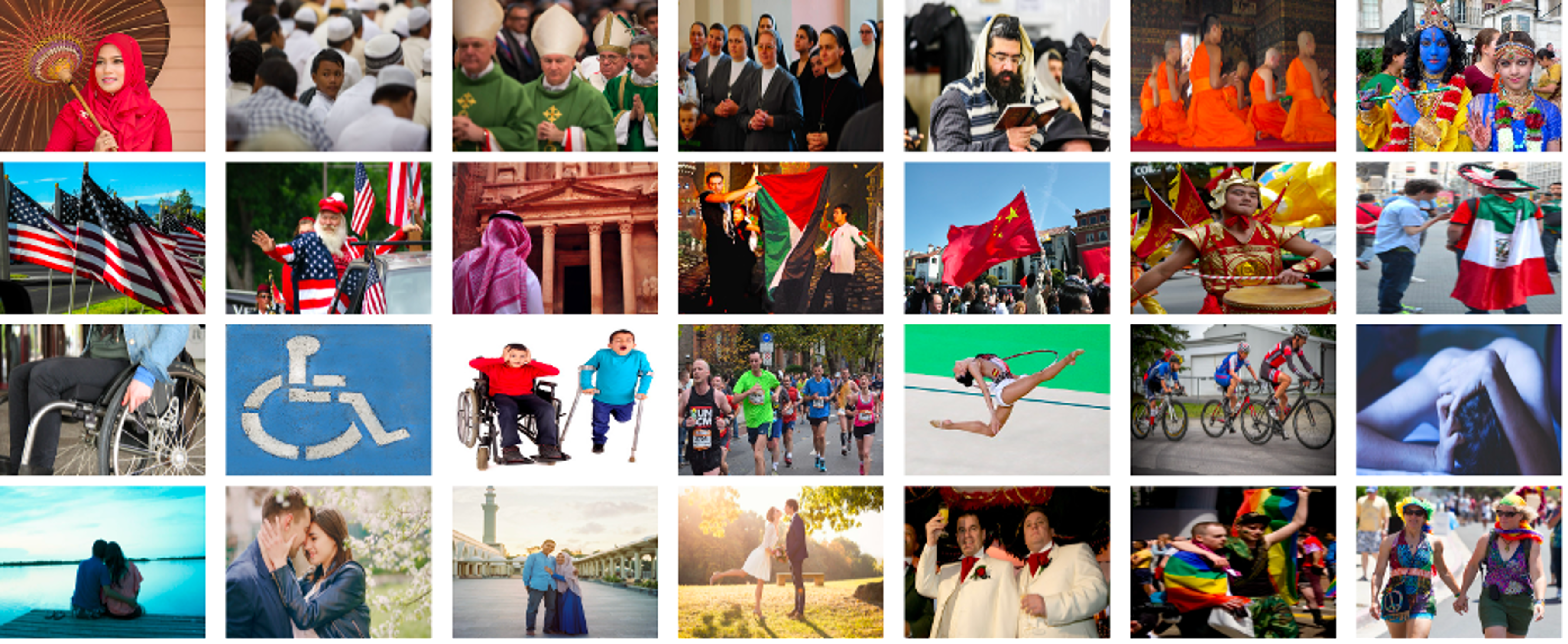}}
	\caption{Sample images from the MMBias dataset. Each row corresponds to one of the target classes: religion, nationality, disability, and sexual orientation. Source: \cite{46}, reproduced with permission.}
	\label{fig1}
\end{figure*}

The risk of bias amplification has been identified in various applications.  Visual Question Answering (VQA), a foundational task in AI, was found to generate incorrect reasoning, highlighting the fragile foundation of these models \citep{25}. In financial earnings calls, gender stereotypes in audio-text models not only perpetuate bias but also ripple into economic outcomes, affecting stock prices \citep{44}. Critical to the health of many, the integration of clinical imaging with demographic data has been found to disproportionately affect underrepresented groups, risking disparities in both predictions and access to care \citep{45}. In line with these applications, an interesting work by \cite{46}, who studied groups vulnerable to bias across nationality, religion, sexual orientation, and disability, demonstrated the propagation and amplification of stereotypes when visual and textual features were combined. Gender bias has often been amplified in fundamental cognitive and affective applications, such as multimodal attention models\citep{39}, emotion recognition \citep{47}, language-image models \citep{48,49}, and audio-visual-embedded stereotypes \citep{33}. Beyond bimodal models, compounded bias effects of verbal, paraverbal, and visual modalities were detected in the hiring processes, further cementing existing stereotypes \citep{31}. Interestingly, even when sensitive attributes, such as gender and ethnicity, were not explicitly included, \cite{32} multimodal models can unearth them from features such as face embeddings and biographies.  Not only do These findings suggest that multimodal models are more prone to errors than their unimodal counterparts, but they also highlight complex dynamics that remain poorly understood, with a high potential to exacerbate inaccurate and unfair predictions and decisions. 

Despite emerging studies addressing bias amplification, significant gaps remain. The primary focus remains on detecting the presence and magnitude of biases, such as using cosine similarity or image-text matching (ITM) probabilities (e.g. \citealp{50,31};), without offering a clear interpretation of the direction of bias interactions. In addition, studies on multimodal interaction or individual modalities have not consistently compared their relative contributions to overall bias (e.g. \citealp{48,47}). In addition, despite few studies highlighting how modalities interact (e.g., \citealp{39,44}) and attempts to compare bias magnitudes across modalities (e.g. \citealp{46}), to my knowledge, no study has quantified these dynamics probabilistically based on modality dominance or classified their interactions dynamically, such as whether biases amplify, mitigate, or neutralize. To fill these significant gaps, this study proposes a systemic, interpretable, and probabilistic framework to assess the dynamic interactions of multimodal biases.  This considerable contribution advances the understanding of multimodal bias interactions and offers a foundation for future research and designing fairer AI systems.

\section{Methodology}

This study introduced a threshold-based comparative system to classify interaction effects, enabling a transparent and intuitive interpretation of interaction dynamics. Leveraging the large MMBias dataset, it evaluates the relative dominance of text bias (St) or image bias (Si) in shaping the multimodal one (Sm), allowing for a nuanced understanding of how individual modalities contribute to, or counteract, biases in multimodal settings. The methodology adopted the following steps:

\subsection{Data Selection:}

This study utilized the MMBias dataset, as introduced and curated by \cite{46} which is freely available under the MIT License and can be accessed via its official repository on GitHub (\url{https://github.com/sepehrjng92/MMBias/blob/main/Readme.md}). This dataset was specifically designed to assess stereotypical bias across vision–language models beyond the traditional categories of gender and race, offering significant information across different target categories.  The dataset encompasses four main target classes, subdivided into 14 specific target groups (Figure \ref{fig1}).

\begin{itemize}
    \item Five major religions ("Muslim," "Christian," "Jewish," "Buddhist," "Hindu");
    \item Four nationalities ("American," "Arab," "Chinese," "Mexican");
    \item Three disability groups ("Mental Disability," "Physical Disability," "Non-disabled");
    \item Two categories addressing sexual orientation ("Heterosexual" and "LGBT").
\end{itemize}

As published in \cite{46}, the dataset has 3,500 target images accompanied by 350 English phrases, —20 phrases per target category—specifically curated to evaluate textual biases. The data provide image-text multimodal associations, as they pair textual phrases with their respective images, facilitating the creation of multimodal embeddings crucial for bias detection experiments.

\subsection{Bias Dynamics’ Definition:}

This study considers that bias in multimodal models could emerge not only from individual modalities (e.g., text or image), but also from their interactions, defining the latter through amplification, mitigation, and neutrality:

\begin{itemize}
    \item Bias amplification: Amplification occurs when the multimodal bias score $Sm$ exceeds the larger of the two unimodal biases $St$ for text-only bias, and $Si$ for image-only bias $Sm>max(St,Si)$. This implies that the interaction between text and image creates a stronger bias than either modality individually.
    \item Bias mitigation: Mitigation occurs when the multimodal bias score $Sm$ is lower than the two unimodal bias scores $Sm<min(St,Si)$. This indicates that the interaction between text and image reduces bias compared to the less biased modality.
    \item Neutral interaction: Neutrality occurs when the multimodal bias score $Sm$ lies between two unimodal bias scores $min(St,Si)\le Sm\le max(St,Si)$. Here, the multimodal interaction neither amplifies nor mitigates bias but reflects or balances one of the modality biases.
\end{itemize}
    
This heuristic approach compares multimodal to individual biases, providing an interpretable and realistic measure of different possible bias interactions. 

\subsection{Bias Score Computation:}

The methodology computes conditional probabilities by leveraging cosine similarity for threshold-based comparisons. This provides statistical insights into how different interaction types—amplification, mitigation, or neutrality—manifest in multimodal systems. The bias scores were computed as follows:

\begin{itemize}
    \item Text-only bias $St$: Derived from the cosine similarity between textual embeddings and sentiment categories (e.g., "pleasant" vs. "unpleasant").
    \item Image-only bias $Si$: Measured by analyzing visual features and their associations with sentiment categories.
    \item Multimodal bias $Sm$: Computed by fusing textual and visual features to capture the overall bias of both modalities.
\end{itemize}

 \subsection{Interaction Classification:}
 
After computing the bias score, the multimodal bias score $Sm$ is compared to the text $St$ and image $Si$ for each subcategory, classifying the interaction into three 3 main categories (Step 2). 

\subsection{Conditional Probabilities: }

To gain deeper insights, this study computed the conditional probabilities of the interaction types given the dominance of one modality over the other, either text dominance as $St>Si$ or image dominance as $Si>St$. These probabilities were calculated as follows:

\begin{multline*}
P(\text{Interaction Type} \mid \text{Modality Dominance}) = \\
\frac{\textit{Count of Interaction Type under Modality Dominance}}{\textit{Total Count under Modality Dominance}}
\end{multline*}

\subsection{Visualization:}

The results were then visualized as bars and conditional probability plots to show bias scores across modalities, interaction effects for relevant categories, and conditional probabilities of interaction types by modality dominance. 

Hence, this methodology not only quantifies bias across individual modalities, but its thresholding also reveals how dynamic bias interactions intensify, alleviate, or balance underlying biases, offering an interpretable and robust tool for understanding bias dynamics in multimodal systems.

\section{Results}

\begin{figure*}[htbp]
	\centering
	\fbox{\includegraphics[width=0.98\textwidth]{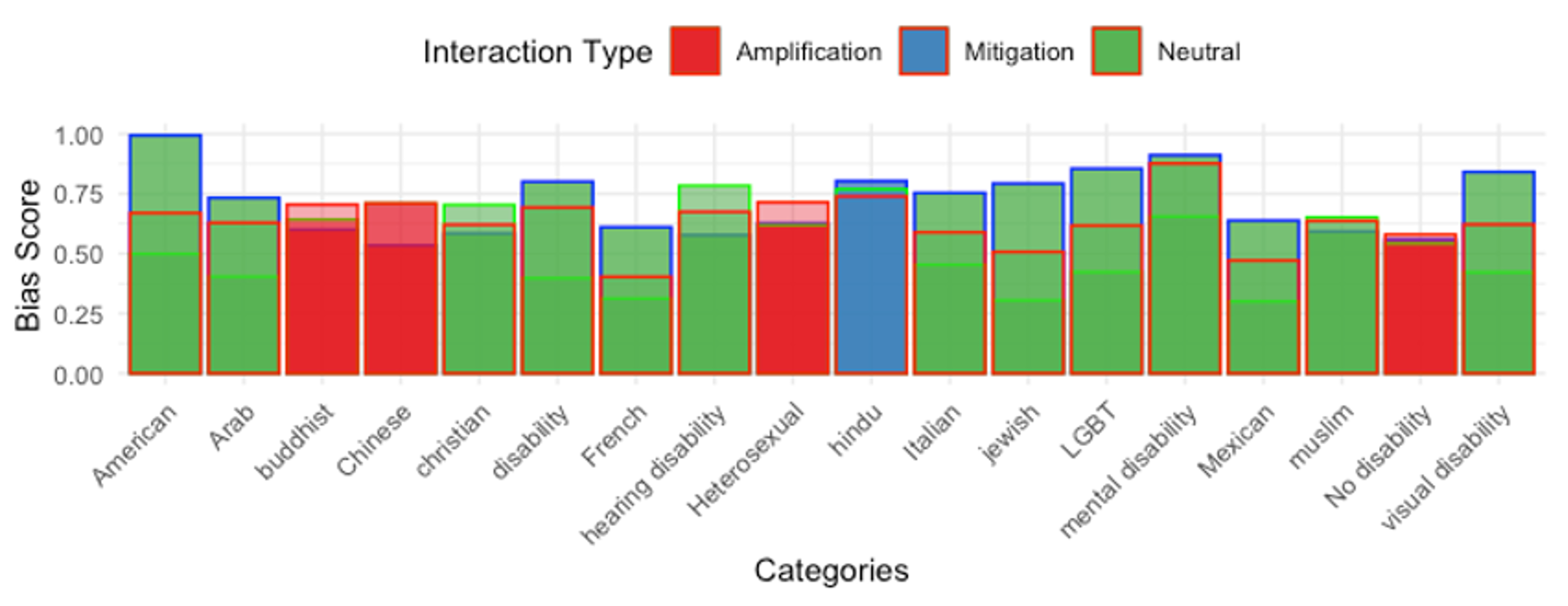}}
	\caption{Bias Scores Across Categories (Author’s compilation)}
	\label{fig2}
\end{figure*}

\subsection{Bias scores:}

The computed bias scores for text, image, and multimodal embeddings revealed significant variability across subcategories, where text bias ($St$) was highest for "American" ($St=0.994$) and lowest for "Chinese" ($St=0.534$);  image bias ($Si$) ranged from $Si=0.300$ ("Mexican") to $Si= 0.784$ ("Hearing Disability"); and multimodal bias ($Sm$) spanned $Sm=0.402$ ("French") to $Sm=0.876$ ("Mental Disability"). Subcategories such as "Mental Disability" and "American" exhibited consistently high bias scores across all modalities, reflecting the inherent biases in both textual and visual inputs.

\subsection{Interaction types:}

The classified interaction effects were observed as follows: For amplification, interactions were found in 4 subcategories (e.g., "Buddhist,” and "Chinese"), suggesting that multimodal models intensified biases when combining textual and visual inputs. Mitigation, however, only occurred in the "Hindu" subcategory, indicating a rare dampening effect. Differently, neutral interaction predominated in 13 subcategories, including "Muslim," "Christian," and "Disability" (Figure \ref{fig2}). 

\subsection{Association Between Interaction and Modalities:}

\begin{figure}[htbp]
    \centering
    \includegraphics[width=\columnwidth]{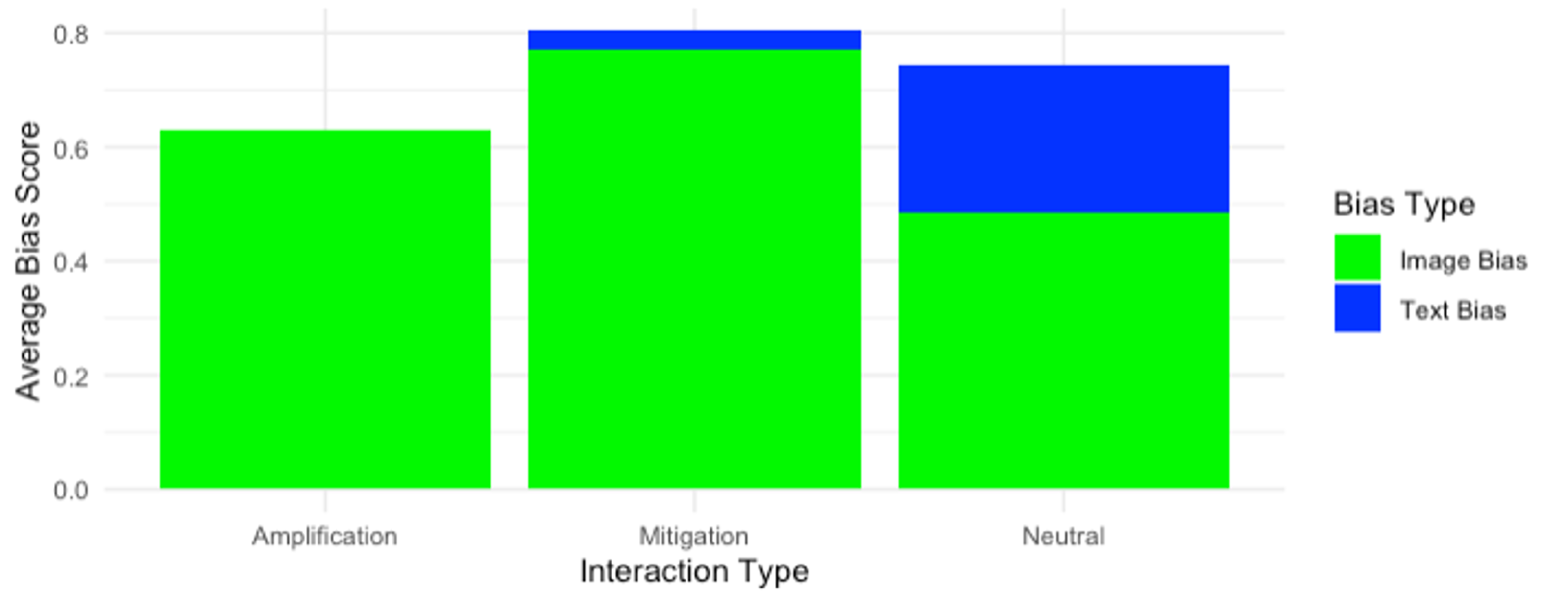}
    \caption{Average Bias Scores by Interaction Type (Author’s compilation).}
    \label{fig3}
\end{figure}

Figure \ref{fig3} highlights how the average bias scores differed across interaction types (amplification, mitigation, and neutral) and the relative contributions of image and text bias. Overall, image scores dominated across all interaction types, suggesting that image content plays a stronger role than text in influencing multimodal bias. This contribution was highest for amplification, indicating that the image may drive increased biases. Interestingly, mitigation exhibited the highest average bias scores, which is counterintuitive because mitigation typically implies a reduction in bias. This may suggest an overlap in the classification thresholds or a less pronounced decrease in multimodal bias. In neutral interactions, the contributions of image and text bias were more balanced, reflecting the combined influence of both modalities rather than dominance by one. 

\begin{figure}[htbp]
    \centering
    \includegraphics[width=\columnwidth]{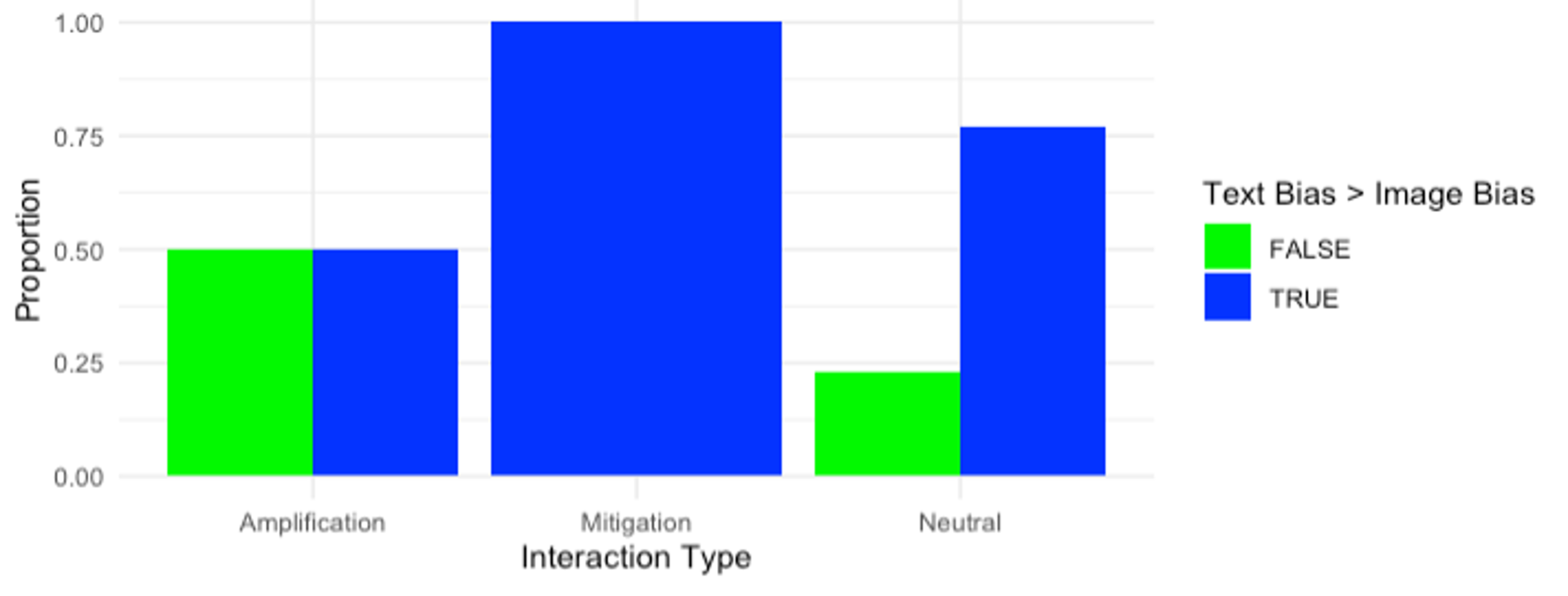}
    \caption{Proportion of Cases by Text vs Image Dominance (Author’s compilation).}
    \label{fig4}
\end{figure}

Conditional probabilities provided greater insight by illustrating the proportion of cases for each interaction type, based on whether text bias exceeded image bias (true/false) (Figure \ref{fig4}). Amplification proportions were split evenly (50\%) between cases where text or image bias dominated, indicating an equal likelihood of either modality contributing more to bias amplification. For mitigation, 100\% of the cases occurred when the bias was higher in the text than in the image, suggesting that it drove most mitigation cases.   Finally, for neutral interactions, image-bias dominance was more prevalent (approximately 77\%) than text-bias dominance (23\%), highlighting the stronger influence of images in maintaining neutral interactions. These observations suggest a highly nuanced dynamic, with amplification occurring with equal likelihood under text or image dominance. Text is crucial in bias mitigation, challenging the notion that neutrality arises more often under image dominance.

\section{Discussion}

In the analyzed dataset, the dominance of image bias across interaction types (Figure \ref{fig3}) raises critical questions regarding the mechanisms by which visual content exerts influence. The pronounced role of images in bias amplification may stem from the ability to evoke stronger associative and emotional responses, aligned with neuropsychological knowledge of visual processing \citep{51}. This suggests that visual information in this dataset, more than textual information, has the potential to reinforce and magnify existing stereotypes, particularly in multimodal systems, where images are given disproportionate weight.

Insights from the conditional probabilities (Figure \ref{fig4}) complement this perspective by focusing on modality dominance in specific interactions. Modality-agnostic bias amplification highlights the need for careful calibration of multimodal fusion mechanisms to avoid unintended magnifications. Conversely, bias mitigation, which occurs only when text bias dominates, demonstrates that text can play a pivotal role in counteracting biases introduced by images. 

The most common interaction type observed in this dataset was neutrality, suggesting that multimodal systems may sometimes have little added value for bias mitigation. For instance, the findings suggest that if image content introduces significant bias, adding text alone is unlikely to mitigate it unless the text is specifically chosen to counteract the bias. This reinforces the recommendation of addressing biases within individual modalities as crucial to preventing their propagation in multimodal systems.

Hence, the insights presented in this study underscore the importance of the suggested probabilistic approach as a first step in assessing biases in the design and evaluation of AI models, ensuring that their integration into societal and organizational contexts promotes fairness. By applying the proposed framework to describe the dynamic interactions of bias in multimodal systems, this study contributes to a deeper understanding of how AI and machine learning can perpetuate societal inequities inadvertently.

\section{Conclusion}

This study provides a systemic framework for classifying and measuring bias amplification, mitigation, and neutrality in multimodal machine-learning models. Through probabilistic simulation and simplistic modelling, and by comparing text, image, and multimodal embeddings across diverse bias-prone categories, the findings reveal nuanced dynamics of bias propagation and underscore the importance of understanding modality-specific contributions to bias to improve fairness in multimodal AI systems. 

Based on these findings, it is highly recommended to integrate this approach into regular audits of single-modality system performance before multimodal integration, calling for a holistic strategy that goes beyond data balancing and integrates bias mitigation in preprocessing, in-processing, and post-processing.   This also includes techniques such as adversarial training, bias-aware loss functions, and the inclusion of diverse and representative training data, along with post-processing methods to adjust biased outputs. 

Future work could also expand this analysis to include additional metrics (e.g., Jensen-Shannon divergence) or extend the use of this heuristic approach to investigate other domains (e.g., audio-video or audio-text embeddings) and task-specific models (e.g., recommendation systems) as bias dynamics could significantly differ due to their functional objectives. Furthermore, the framework can be expanded to include different multimodal model architectures. For instance, early fusion models that merge features at the input level could introduce compounded biases early on from both modalities, whereas generative models such as DALL-Es and GPTs, actively generating outputs by interpreting cross-modal information, can introduce emergent biases that are not present in individual modalities.  Hence, future research could explore how multimodal architectural design and fusion strategies can affect interaction dynamics. 

Overall, further research is required to explore the conditions under which bias mitigation is most effective, recommending the use of this framework and cross-validation with alternative measurement tools and real-world applications for more robust and actionable recommendations.  As multimodal AI systems have become more prevalent, understanding and addressing the complex interactions of biases across modalities is imperative. By implementing thoughtful design and mitigation strategies, we can harness the full potential of these systems, while promoting fairness, equity, and accuracy in their predictions and applications.

\nocite{*}
\bibliographystyle{unsrtnat}
\bibliography{main}  

\end{document}